\documentclass[runningheads]{llncs}
\usepackage{graphicx}

\usepackage{tabularray}
\usepackage{multirow}
\usepackage{bm}
\usepackage{amsmath}
\usepackage{amssymb}
\usepackage{booktabs}
\usepackage{bbding}
\usepackage{tablefootnote}
\def\eg{\emph{e.g.}}
\def\ie{\emph{i.e.}}
\usepackage{hyperref}
\begin{document}
\title{Centroid-centered Modeling for Efficient Vision Transformer Pre-training}
%
%
\author{Xin Yan \and
Zuchao Li\thanks{This work was supported by the
National Natural Science Foundation of China (No. 62306216) and the Natural Science Foundation of Hubei Province of China (No. 2023AFB816).} \and
Lefei Zhang}
\authorrunning{X. Yan et al.}
%
\institute{Wuhan University, Wuhan, China \\ \email{zcli.charlie@gmail.com}}
%
\maketitle              
\begin{abstract}
Masked Image Modeling (MIM) is a new self-supervised vision pre-training paradigm using a Vision Transformer (ViT). Previous works can be pixel-based or token-based, using original pixels or discrete visual tokens from parametric tokenizer models, respectively. Our proposed centroid-based approach, \textbf{CCViT}, leverages k-means clustering to obtain centroids for image modeling without supervised training of the tokenizer model, which only takes seconds to create. This non-parametric centroid tokenizer only takes seconds to create and is faster for token inference. The centroids can represent both patch pixels and index tokens with the property of local invariance. Specifically, we adopt patch masking and centroid replacing strategies to construct corrupted inputs, and two stacked encoder blocks to predict corrupted patch tokens and reconstruct original patch pixels. Experiments show that our \textbf{CCViT} achieves 84.4\% top-1 accuracy on ImageNet-1K classification with ViT-B and 86.0\% with ViT-L. We also transfer our pre-trained model to other downstream tasks. Our approach achieves competitive results with recent baselines without external supervision and distillation training from other models.\footnote{Codes are available at \href{https://github.com/Cakeyan/CCViT_Public}{https://github.com/Cakeyan/CCViT\_Public}}

\end{abstract}

\section{Introduction}

Over the past several years, Vision Transformer (ViT)\cite{dosovitskiy2020image} has been verified to produce exceptional results in image modeling and other tasks. However, empirical studies have shown that Vision Transformer requires a larger volume of data than previous CNN-based models. 

The issue has been successfully solved by self-supervised learning. Particularly in NLP, BERT\cite{devlin2018bert} has proposed Masked Language Modeling (MLM) to solve the problem. Such mask-then-predict task masks out some proportion of the input data and then learns to predict the masked target, showing its superiority in leveraging large-scale unlabeled data. Inspired by MLM, multiple studies have attempted to introduce the task into computer vision. Regarding discrete visual tokens as targets, BEiT\cite{bao2022beit} proposed Masked Image Modeling (MIM), using a tokenizer to convert an image into tokens, which is obtained by an extra training stage with a decoder via discrete variational autoencoder (dVAE)\cite{ramesh2021zero} method. During pre-training, BEiT takes corrupted images as input and learns to recover the masked tokens. 
Contrarily, MAE\cite{he2022masked} reconstructs raw image pixels via an additional decoder. 

\begin{figure}[t]
    \begin{center}
       \includegraphics[width=0.95\linewidth]{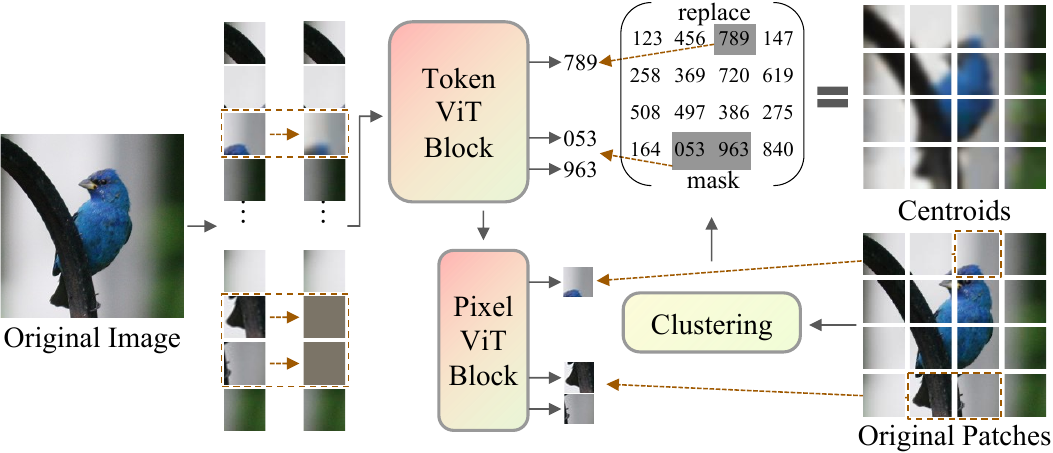}
    \end{center}
    \caption{\textbf{The proposed CCViT architecture.} We view centroids as two aspects, token ids, and patch pixels. Our centroid-centered pre-training aims at both predicting centroid indices and reconstructing image patch pixels. We apply the blockwise mask to some patches (\eg, 40\%) and replace some of the remaining patches (\eg, 10\%) with the corresponding centroids. All corrupted patches are fed into the ViT Block. } 
    \label{fig:model_small}
\end{figure}

In contrast to primarily language tokens as targets in NLP, various reconstruction targets have emerged in computer vision, including visual tokens\cite{li2022mc,Dong2021perceptual}, high-level features\cite{chen2022sdae}, vanilla pixels\cite{he2022masked} and original image features\cite{wei2022masked}, due to different information density between vision and language. 
We can broadly categorize these targets into two types of models: token-based MIM with tokens or high-level features as the pre-training objective, which typically requires an additional training stage of a parametric tokenizer model for image abstraction from continuous to discrete; and pixel-based MIM, which encourages the model to reconstruct raw pixels or original features such as HOG without tokenizers. 

However, both of these two methods have their drawbacks. 
First, they both introduce redundant parametric modules to convert latent representations into raw pixels or visual tokens, while discarding them when fine-tuning. These modules need an additional training stage or extension of pre-training consumption. More importantly, they may introduce additional datasets or strong supervision.
Moreover, each visual token from parametric tokenizers cannot represent the corresponding image patch, while the actual image and mask-then-predict framework contradict this, which relies on this correlation between tokens for masking inference.
This also accounts for achieving high token prediction accuracy in pre-training is typically challenging in token-based MIM. 
We will provide a comprehensive proof in the Experiments Section and Table~\ref{table: cmp_tokenizer}.

In this work, we introduce \textbf{CCViT}, which stands for \textbf{C}entroid-\textbf{c}entered \textbf{Vi}sion \textbf{T}ransformer, as shown in Figure~\ref{fig:model_small}. Specifically for our centroid-based tokenizer, we utilize an efficient clustering method (such as the k-means algorithm) on vanilla patch pixels to identify the token id of each patch, which is the index of the nearest centroid. 
Different from parametric tokenizers consuming large training resources before pre-training, the centroids can be easily obtained using a very small proportion of training set with only a few seconds. Also, it is faster to obtain visual tokens compared with other parametric tokenizers. More importantly, we only perform clustering on the training set of ImageNet-1K\cite{deng2009imagenet} to obtain centroids, without potentially introducing large or private datasets such as DALL-E\cite{bao2022beit,ramesh2021zero}, or distilling from large models such as CLIP\cite{peng2022beit,radford2021learning} into the tokenizer. During pre-training, we split images into several patches and masked out some proportion of image patches. We feed all of the corrupted patches into our backbone ViT encoder to reconstruct the centroid indices of these patches.

We also propose a novel perspective of MIM, namely, centroid-based MIM, which streamlines the process and eliminates the need for additional training costs. 
In our centroid-based MIM, we focus on the clustering centroids, which can be considered both centroid patch pixels and centroid index tokens. To fully leverage this dual-natured property, we adopt a random replacing masking strategy that will replace some image patches with the nearest centroid pixels to encourage the model to align the pixel representations of centroids with their corresponding token ids. 
We also stack two ViT blocks with different learning objectives, including a token ViT block and a pixel ViT block. 
Further experiments and analysis have demonstrated that these two strategies effectively utilize this dual nature and improve performance.

We conduct self-supervised pre-training on ImageNet-1K and fine-tune our pre-trained model on downstream tasks, \ie image classification and semantic segmentation. Experimental results show that our method achieves excellent performance for image representation. 
Concretely, we achieve 84.3\% on ImageNet-1K classification and 48.4 on ADE20K mIoU with ViT-B, outperforming BEiT by +1.4\% and +3.7 with the same 300 pre-training epochs. 
To demonstrate the scaling ability, we adopt larger backbones and longer pre-training epochs, and achieve notable improvement.
Ablation studies show learning together for both sides of centroids, pixels, and tokens, performs better than learning singularly of either form.
Besides, we analyze why our centroid-based MIM is more suitable to the mask-then-predict framework and has the highest token prediction accuracy. The contributions of this work can be summarized as follows:

\begin{itemize}
    \item We propose a novel non-parametric visual tokenizer based on k-means clustering. This tokenizer is rapid to create and only requires a small proportion of the training set, eliminating the need for external data. We also introduce a novel random replacing strategy and a stacked model architecture.

    \item We categorize all existing MIM methods by their pre-training objectives into token-based MIM and pixel-based MIM and propose our centroid-based MIM, offering a dual perspective by treating centroids as both patch pixels and index tokens. This approach requires a lower tokenizer training cost than token-based MIM, and is more computationally efficient by removing the decoder compared to pixel-based MIM.

    \item We conduct extensive experiments on downstream tasks and transfer tasks to demonstrate our state-of-the-art performance without external supervision and distillation. Additionally, we conduct an efficiency analysis including FLOPs, model parameters, and so on.

    \item We thoroughly investigate different visual tokenizers and find that previous visual tokens cannot represent corresponding image patches, compared with our dual-natured centroids. Experiments show that our centroid tokenizer exhibits robust performance and can be extended to other tasks including pixel reconstruction.

\end{itemize}

\section{Related Work}

\noindent\textbf{Self-supervised visual learning} has been explored over the years to introduce such a learning paradigm into vision pre-training. Various methods use different pre-text tasks for pre-training, including jigsaw puzzle\cite{Noroozi2016Unsupervised}, colorization\cite{Larsson2017Colorization} and predicting rotation\cite{gidaris2018unsupervised}. Contrastive learning is also a trend in visual representation learning\cite{chen2020simple,chen2021exploring,grill2020bootstrap,he2020momentum,wu2018unsupervised}. These methods typically rely on data augmentation approaches. Also, early studies perform clustering methods to learn vision representation\cite{caron2018deep,caron2020unsupervised,li2021prototypical,YM2020self}. Most recently, iGPT\cite{chen2020generative} creates a 9-bit color palette by clustering (R, G, B) pixel values using k-means with $K = 512$ and uses the clustered token sequence as the direct input of both auto-regressive and BERT objectives. 
In comparison, our method uses image patches as the input, and centroid indices as the pre-training objective. 

\noindent\textbf{Masked image modeling}  has seen widespread application in the field of visual pre-training as a counterpart of masked language modeling method in NLP, \eg, {BERT}\cite{devlin2018bert}. Since ViT\cite{dosovitskiy2020image} first outcome the obstacle of architecture, masked image modeling (MIM) has achieved remarkable success rapidly\cite{caron2021emerging,chen2022context,el2021large}. MIM randomly masks some proportion of an image and reconstructs it in the pre-training stage. Due to different reconstruction targets in the pre-training stage rather than primarily tokens as an objective in NLP, there are two mainstream paradigms in MIM methods, \ie, token-based MIM and pixel-based MIM. 
Token-based MIM derives from the prior art, BEiT\cite{bao2022beit}, and needs a parametric tokenizer to generate tokens or high-level features for pre-training targets. To be specific, BEiT and VIMPAC\cite{tan2021vimpac} use an offline discrete VAE tokenizer from DALL-E\cite{ramesh2021zero,rolfe2017discrete}. PeCo\cite{Dong2021perceptual} regards MoCov3\cite{chen2021empirical} as the perceptual model in VQGAN\cite{esser2021taming} training. mc-BEiT\cite{li2022mc} also focuses on perceptual similarity and constructs a codebook for pre-training. iBOT\cite{zhou2021ibot}, SdAE\cite{chen2022sdae} and data2vec\cite{baevski2022data2vec} use self-distillation method.  BEiTv2\cite{peng2022beit} also utilizes distillation method and introduces vector-quantized knowledge distillation\cite{van2017neural} to train the tokenizer using CLIP\cite{radford2021learning}. 
Pixel-based MIM with non-parametric tokenizers such as MAE\cite{he2022masked} and SplitMask\cite{el2021large}, considers vanilla pixels or patches as pre-training targets instead of tokens and need a redundant decoder. 
MaskFeat\cite{wei2022masked} further introduces hand-crafted HOG features as the targets. Different from these works, our centroid-based MIM uses a non-parametric tokenizer to model both tokens and pixels in pre-training. We use a k-means clustering algorithm to generate centroids which does not suffer from the costs for tokenizer model training.

\section{Masked Image Modeling}

MIM, which was first proposed in BEiT, introduced BERT-style pre-training into computer vision and has successfully replicated the success of NLP. Specifically, it first splits an input 2D image into a sequence of patches, and then masks out a proportion of the patches. The pre-training objective of token-based MIM is reconstructing corrupted images using the visual context at a higher semantic level,  \ie, discrete visual tokens, compared to pixels. Formally, given an input image $x \in \mathbb{R}^{C\times H\times W}$, it is first split into $n=HW/P^2$ patches and flattened to $\{x^p_i\}^n_{i=1}$, where $(P,P)$ is the resolution of a patch and $x^p \in \mathbb{R}^{n\times{C\times P^2}}$. It samples a mask via the mask ratio $r_m$ on input image $x$ to generate the corrupted image $\hat{x}$ according to the masked positions ${\mathcal{M}}$: 
\begin{equation}
    \hat{x}=\{x^p_i\boldsymbol{E}_p \mid i\notin \mathcal{M} \}^n_{i=1} \bigcup \{e_m \mid i\in \mathcal{M}\}^n_{i=1}
\end{equation}
where $e_m$ is the learnable mask token embedding, $x^p_i\boldsymbol{E}_p$ indicates the process of patch embedding calculation.

The objective of pre-training is to recover visual tokens $\{t_i\}^n_i$ obtained by the tokenizer, where $t\in \mathcal{V}^n$, $\mathcal{V}=\{1,\ldots K\}$ is the $K$-size codebook with discrete token ids. The corrupted image will be encoded into $\psi(\hat{x}) \in \mathbb{R}^{n\times d}$ through the ViT encoder. The representation will be fed into a linear classification head $lin.: \mathbb{R}^d \to \mathbb{R}^K $ and a softmax operator to obtain the probabilities $p_{\mathrm{MIM}}(t_i\mid\hat{x})=\mathrm{softmax}_{t_i}(lin.\circ\psi(\hat{x}))$ of each token. The model is learned by the cross-entropy loss, which is to maximize the log-likelihood of correct tokens $t_i$ via corrupted image $\hat{x}$: 
\begin{equation}
\max \sum_{x} \mathbb{E}_{\mathcal{M}}\left[\sum_{i \in \mathcal{M}} \log p_{\mathrm{MIM}}\left(t_i \mid \hat{x}\right)\right]    
\end{equation}
where $p_{\mathrm{MIM}}\left(t_i \mid \hat{x}\right)$ is the softmax probability after feeding $\hat{x}$ into the ViT encoder to predict the correct tokens $t_i$.

\begin{figure*}
    \begin{center}
       \includegraphics[width=0.99\linewidth]{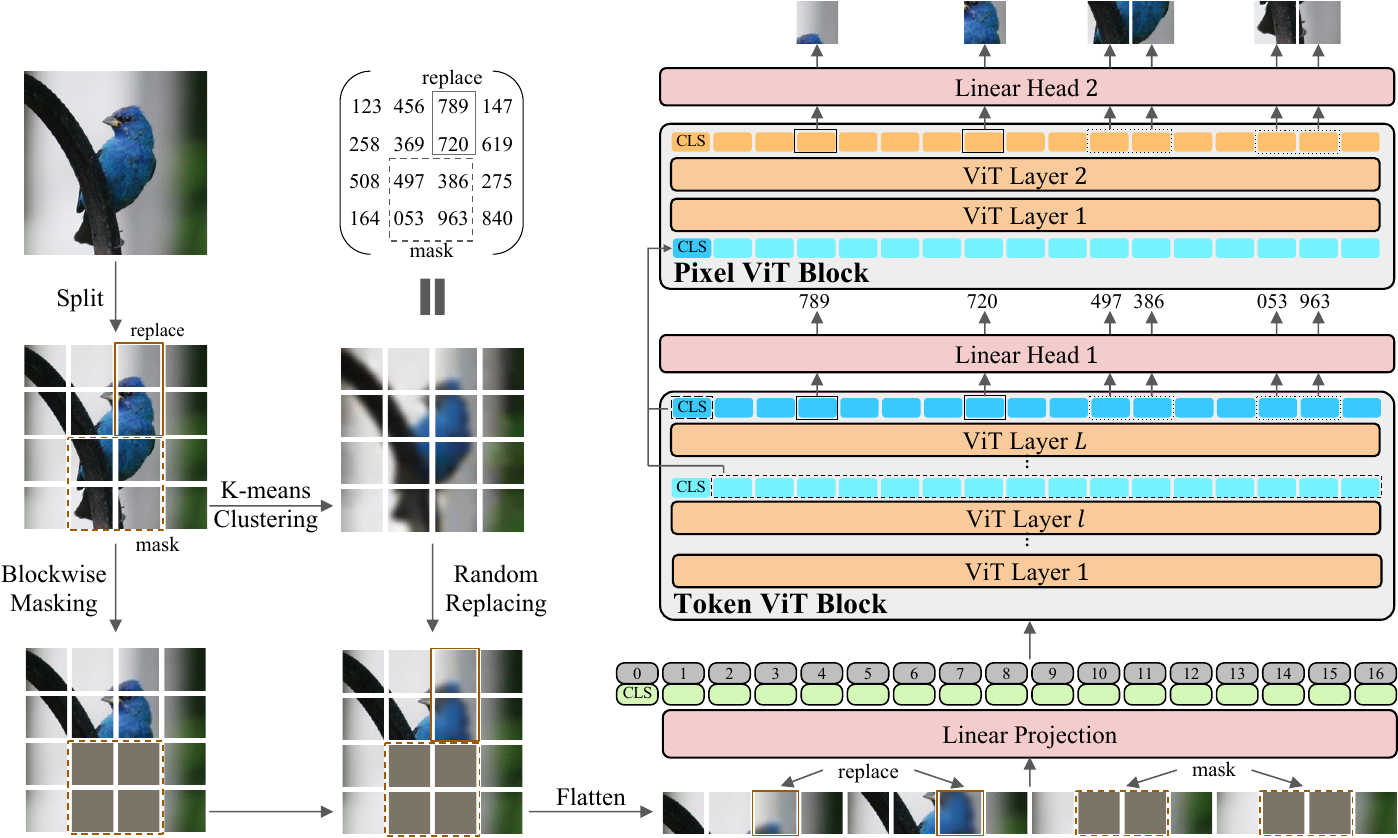}
    \end{center}
    \vspace{-10pt}
    \caption{Overview of our CCViT. Before pre-training, we use k-means clustering on vanilla pixel patches to achieve centroids. During pre-training, we mask out some patches and randomly replace some of the remained patches using centroids. All the patches are fed into the encoder. The pre-training objectives are both centroid index tokens and original pixels.}
    \label{fig:model_big}
    \vspace{-10pt}
\end{figure*}

\section{Our Method}
In our CCViT, we propose centroid-based MIM, which uses clustering centroids to model the images. The whole architecture is shown in Figure~\ref{fig:model_big}. The centroids have two views, \ie, pixel view and token view. We can directly obtain the pixels as each centroid is like an averaged image patch. We can also easily obtain the tokens which are the indices of centroids, as each centroid patch corresponds to its index. It is for the dual-natured property, the output can be either tokens or pixels with different loss functions, which are cross-entropy loss and mean squared error respectively.
 We feed all the patches into the backbone ViT encoder, including original patches, masked patches, and replaced patches with centroids. 

\subsection{Centroid-based Tokenizer}
Token-based MIM needs to convert the continuous images to discrete tokens first to pre-train the model with a patch classification objective implemented with a cross-entropy loss.
It usually employs a parametric tokenizer model trained on a large scale of the dataset.
Our centroid-based tokenizer does not require extensive resources to train an additional parametric tokenizer model. We only utilize clustering methods, such as the k-means algorithm, to obtain a set of centroids. The quality of the centroids via the k-means algorithm does not heavily rely on a high standard and large scales for images and can be achieved by only utilizing a small proportion (\eg\ 4\% of ImageNet) of the training set used for pre-training, thereby circumventing the risk of introducing implicit additional datasets that often occurs in parametric tokenizer models in token-based MIM. 

We use the k-means algorithm to partition each vanilla image patch into $K$ clusters. Given the patches $\{x^p_i\}^n_{i=1}$, they will be flattened into a vector with the dimension of $D=C\times P^2$. So $N$ $D$-dimensional vectors $\mathcal{X}=\{x^v_i \in\mathbb{R}^D\}_{i=1}^N$ will find $k$ centroids $\{{\mathcal{C}}_k \in \mathbb{R}^D\}_{k=1}^K$ that minimize the cost:
\begin{equation}
\mathbb{E}({\mathcal{C}}_1, \ldots, {\mathcal{C}}_K)=\frac{1}{N} \sum_{i=1}^N \|x^v_i- {\mathcal{C}}_{a(i)}\|_2
\end{equation}
where $N$ is the number of training images of k-means cluster and $a(i)$ is an assignment function and is defined by $a(i)=\mathop{\arg\min}_{k \in\{1, \ldots, K\}} \|x^v_i - \mathcal{C}_{k}\|_2$. Our centroid-based tokenizer has the property of local invariance as we only operate on each patch separately in a local perspective. 

In the process of converting continuous patches to discrete visual tokens, for image patch $x^p_i \in \mathbb{R}^{C\times P^2}$ and its flattened vector $x^v_i \in \mathbb{R}^D$, its index $t_i \in \{1,\ldots K\}$ can be obtained by looking up the index of nearest centroid, which is:
$t_i=\mathop{\arg\min}_{k \in\{1, \ldots, K\}} \|x^v_i - \mathcal{C}_{k}\|_2$

Notably, as we only do clustering operations and do not change the dimension of each patch, the centroid itself is a patch that can be easily visualized. This enables our centroid-based non-parametric tokenizer to have the dual-natured property that can be seen as both pixels and tokens. In other words, we can get an approximate version of the input patches based on the token ids.

\subsection{Image Corruption For Modeling}

Self-supervised pre-training is a process of corrupting the image and then using internal correlations of the image to reconstruct the original image or other forms. In our CCViT,  we employed two corruption strategies: blockwise masking and centroid replacing.
Blockwise masking is more challenging compared with random masking as it provides less visual context. 
In centroid replacing, we randomly replace some proportion of the remained patches with the corresponding centroids. Such a technique allows the model to learn the mapping relationships between centroids and indices, centroid approximate pixels, and raw pixels. This makes our model more noise-resistant and thus achieves better results.

Formally, given masking ratio $r_m$ and replacing ratio $r_{re}$, the masked positions can be obtained as ${\mathcal{M}} \in \{1, \ldots, n\}^{r_m \times n}$ and the replaced positions are ${\mathcal{R}} \in \{1, \ldots, n\}^{r_{re} \times n}$. Notice that the replaced positions are always the unmasked positions to avoid being masked, \ie, $\mathcal{M} \bigcap \mathcal{R} = \emptyset$. So an input image $x \in \mathbb{R}^{C\times H\times W}$ and its patch series $\{x^p_i\}^n_{i=1}$ can be corrupted into $\Tilde{x}$:
\begin{equation}
\begin{aligned}
\Tilde{x}=\{x^p_i \boldsymbol{E}_p \mid i\notin \mathcal{M} \bigcup \mathcal{R} \}^n_{i=1} \bigcup \{e_m \mid i\in \mathcal{M}\}^n_{i=1} \bigcup \{C_{a(i)} \boldsymbol{E}_p \mid i\in \mathcal{R} \}^n_{i=1}
\end{aligned}
\end{equation}
where $e_m$ is the learnable mask token embedding and $\mathcal{C}_i$ is the centroid. And $a(i)$ is the assignment function and is defined by $a(i)=\mathop{\arg\min}_{k \in\{1, \ldots, K\}} \|x^v_i - \mathcal{C}_{k}\|_2$.

\subsection{Vision Transformer Backbone}

We use ViT as our backbone network as illustrated in Figure~\ref{fig:model_big}. The input is the corrupted image patches $\Tilde{x}$ with the corrupted positions $\mathcal{T}=\mathcal{M}\bigcup\mathcal{R}$, and is flattened and mapped to $d$-dimensional patch embeddings $\Tilde{x_i}\boldsymbol{E}_p$, where $\boldsymbol{E}_p\in \mathbb{R}^{d\times (C \times P^2)}$ is a mapping projection of a single 2D convolutional layer. 

Due to the local perspective of our centroid-based modeling, we need the {\tt CLS} token to aggregate the global representation explicitly to compensate for the semantic damage caused by downsampling original image patches to centroids. We add pixel ViT block to utilize {\tt CLS} token to gather global information, which was effective in prior token-based MIM\cite{gao2021condenser,peng2022beit}. 
The input vectors $\boldsymbol{H}_0=[\boldsymbol{E}_{\tt CLS}, \Tilde{x_1}\boldsymbol{E}_p,\ldots,\Tilde{x_n}\boldsymbol{E}_p]$ will pass through $L$-layers token ViT block $\psi^L_t(\Tilde{x})$ to predict masked and replaced tokens. 
We concatenate the early representations from the $l$-th layer $[\boldsymbol{h}_1^l,\ldots,\boldsymbol{h}_n^l]$ and the {\tt CLS} token from the $L$-th layer $\boldsymbol{h}^L_{\tt CLS}$.

The concatenated vector $\boldsymbol{H}_{p} = [\boldsymbol{h}^L_{\tt CLS},\boldsymbol{h}_1^l,\ldots,\boldsymbol{h}_n^l]$
 is used as input of pixel ViT block for raw pixels reconstruction.
The pixel block $\psi^{2}_p(\boldsymbol{H}_p)$ is composed of only two layers since the token form of centroids is closely related to the pixel form. 
To obtain the predicted tokens and reconstructed image, the representations from the token block and pixel block will be mapped into centroid index space and original pixel space using two different linear heads $lin.1(\cdot)$ and $lin.2(\cdot)$. 

Two losses are thus computed, and the cross-entropy loss is utilized to measure the likelihood between the ground-truth centroid indices and predicted index probabilities $p_{\mathrm{CIM}}(t_i\mid\Tilde{x})=\mathrm{softmax}_{t_i}(lin.1\circ\psi^L_t(\Tilde{x}))$, which is the output of token block. The Mean Square Error (MSE) is used to quantify the dissimilarities between the image patches generated by pixel block $\psi^{2}_p(\boldsymbol{H}_p)^\mathcal{T}$, and the original image patches ${x^p}^{\mathcal{T}}$. 
The total loss of our CCViT pre-training $\mathcal{L}_{\mathrm{CIM}} = \mathcal{L}_{\mathrm{CE}} +\mathcal{L}_{\mathrm{MSE}}$ can be formulated as:
\begin{equation}
\mathcal{L}_{\mathrm{CE}} + \mathcal{L}_{\mathrm{MSE}} = - \sum_{\Tilde{x}} \sum_{i\in \mathcal{T}}(\log( p_{\mathrm{CIM}}\left(t_i \mid \Tilde{x}\right)) - \frac{1}{\lambda(\Tilde{x}^\mathcal{T})}\|{x^p}^{\mathcal{T}}-\psi^{2}_p(\boldsymbol{H}_p)^\mathcal{T}\|_2)
\end{equation}
where $\lambda(\Tilde{x}^\mathcal{T})$ is the number of elements and $\psi^{2}_p(\boldsymbol{H}_p)^\mathcal{T}$ means that we input the entire corrupted patches (masked, replaced and original patches) and only compute the loss on the corrupted portion (masked and replaced patches).

\section{Experiments}

We perform experiments on downstream tasks using our pre-trained model, including image classification and semantic segmentation, following the standardized evaluation protocols that have been adopted in previous works\cite{bao2022beit}. We also conduct a brief ablation study on critical components of our model. 

\subsection{Pre-training Setup}

\paragraph{Centroids Clustering.} We adopt the clustering method on each vanilla pixel patch to get the centroids and their corresponding index. We use the Faiss-GPU\cite{johnson2019billion} library for the k-means algorithm and set the number of clusters $K=8192$ for a fair comparison to existing works.
We use a small proportion (50K images) of ImageNet-1K training set for training centroids as it is the dataset used in pre-training, thus will not implicitly introduce the information from other datasets. 
In our experiment, each image is resized to $224\times 224$ resolution and will be split into $14\times 14$ image patches. So each patch has the feature size of $d=3\times 16\times 16=768$. 
According to the experience of conducting k-means clustering on samples from Faiss, there is no consistent improvement of the k-means quantizer beyond 20 iterations and $1000 \times K$ training points. So we randomly chose 50 images in each class of ImageNet-1K training set to train the centroids with 20 iterations, namely 50K images and 9.8M image patches which is only 4\% of it. Such a clustering stage will only need about 150 seconds, which is a significant improvement compared with days of parametric tokenizer model training.

\paragraph{Centroid-based MIM.} 
We follow the settings for fair comparison used in MIM methods\cite{bao2022beit,peng2022beit,li2022mc}. We use the ImageNet-1K training set without labels to pre-train our model via self-supervised learning. We set the resolution of an image as $224\times 224$ patch size as $16\times 16$, and pre-train base-size Vision Transformer (ViT-Base/16). We use block-wise masking with a ratio of 40\% and replaced centroids with 10\%, which is 75 patches and 20 patches of the remaining respectively. The base-size ViT has $L=12$ layers in the token block and we append two layers of pixel block whose input is the output of the $l=9$-th layer concatenated with the {\tt CLS} token from the last $L=12$-th layer of the token block. 
We further demonstrate the effectiveness of our approach when scaling to ViT-Large.

We pre-train our model for 300 epochs using 4 NVIDIA GeForce RTX 3090 GPUs. 
To keep the same batch size as existing works, we accumulate the gradients by 4 times resulting in a 2048 total batch size. 
More hyperparameters can be found in the Appendix.

\begin{table}[hbt]
    \small
    \centering
    \caption{Main results comparison of our model and previous works. Fine-tuning Results are mainly based on the ViT-Base model on downstream tasks, including top-1 accuracy (\%) of image classification on ImageNet-1K and mIoU (\%) of semantic segmentation on ADE20K. $^\dag$: result reproduced by MAE. $^\ddag$: Non-parametric methods, whose tokenizers have no parameters.}
    \label{table:results}
    \begin{tabular}{lcccccc}
    \toprule
    \bf Method   & \bf Supervision  & \bf Non-p.$^\ddag$ & \bf External Data & \bf Epoch & \bf Acc & \bf mIoU        \\
    \midrule
    Supervised ViT\cite{he2022masked}    & Label     & \CheckmarkBold       &      \XSolidBrush         & -     & 77.9     & 47.4$^\dag$        \\
    \midrule
    \multicolumn{7}{l}{\textit{Token-based with parametric tokenizer}} \\
    BEiT\cite{bao2022beit}    & DALL-E    & \XSolidBrush        &      \CheckmarkBold               & 300   & 82.9     & 44.7        \\
    mc-BEiT\cite{li2022mc} & VQGAN       & \XSolidBrush       &      \XSolidBrush     & 300   & 83.9     & -           \\
    iBOT\cite{zhou2021ibot}    & Self-Distillation    & \XSolidBrush         &      \XSolidBrush            & 1600  & 84.0     & 50.0        \\
    PeCo\cite{Dong2021perceptual}& VQGAN+MoCo  & \XSolidBrush   &      \CheckmarkBold & 300   & 84.1     & 46.7        \\
    BEiTv2\cite{peng2022beit}    & CLIP-B      & \XSolidBrush          &      \CheckmarkBold          & 300   & 85.0     & 52.7        \\
    \midrule
    \multicolumn{7}{l}{\textit{Pixel-based}} \\
    MAE\cite{he2022masked}        & Pixel     & \CheckmarkBold       &      \XSolidBrush              & 1600  & 83.6     & 48.1        \\
    SplitMask\cite{el2021large}    & Patch     & \CheckmarkBold      &      \XSolidBrush                 & 300  & 83.6     & 45.7        \\
    MaskFeat\cite{wei2022masked}   & HOG     & \CheckmarkBold       &      \XSolidBrush                & 300   & 83.6     & -           \\
    RILS\cite{yang2023rils}        & Pixel+LAION     & \CheckmarkBold       &      \CheckmarkBold              & 1000  & 83.6     & 48.1        \\
    \midrule
    \multicolumn{7}{l}{\textit{Centroid-based}} \\
    \bf Ours     &\bf  Centroids   & \CheckmarkBold &      \XSolidBrush      & \bf  300   & \bf 84.3     & \bf  48.4   \\
    \bottomrule
    \end{tabular}
\end{table}

\begin{table}[ht]
    \centering
    \small
    \caption{Fine-tuning results of scaling ability on ImageNet-1K. We present larger backbones (ViT-L) and longer epochs (800 epochs). }
    \label{table:vit-l}
    \begin{tabular}{lcccc}
    \toprule
    \bf Method    & \bf Supervision & \bf ViT-B & \bf +Longer & \bf +Larger       \\
    \midrule
    mc-BEiT      & VQGAN          & 83.9 &    84.1    & 85.6 \\
    iBOT      & Distillation        &  83.8 &    84.0   & 84.8 \\
    PeCo      & MoCo                      & 84.1 &  84.5     & 86.5 \\
    BEiTv2      & CLIP-B             & 85.0  & -     & 86.6 \\
    MaskFeat   & HOG                & 83.6 & 83.9   & 84.4 \\
    BEiT      & DALL-E       & 82.9 & 83.2       & 85.2 \\
    MAE   & Pixel              & -   & 83.6  & 85.9 \\
    
    \midrule
    \bf Ours      &  \bf Centroids & \bf 84.3  &   \bf 84.4  & \bf 86.0   \\
    \bottomrule
    \end{tabular}
\end{table}

\subsection{Image Classification}

Image Classification on ImageNet-1K is the main task of visual pre-training, as the dataset is the same as the pre-training dataset.
For a fair comparison, we only use the token ViT block module in fine-tuning to ensure we have the same size of ViT model parameters.
We adopt top-1 accuracy after fine-tuning for 100 epochs as the metric. 
More hyperparameters can be found in the Appendix.

We present the image classification results of ViT-B in Table~\ref{table:results}. Our base-size model only is pre-trained for 300 epochs and reaches 84.3\% top-1 accuracy on ImageNet-1K classification, which outperforms the MIM baseline methods BEiT by +1.4\%, MAE by +0.7\%. 
This verifies the effectiveness of our CCViT with centroid-based MIM as we only use the ImageNet-1K dataset and an easily constructed supervision with no parameters. 

Recent works often introduce potential external datasets or strong supervision into models, including DALL-E and CLIP. Baselines equipped with external information often achieve significant improvements. However, it is not a fair assessment to compare with others without external data. We outperform all the baselines with ImageNet-1K only data, and even baselines with external data such as BEiT and PeCo. 
Our model performs worse than BEiTv2, which utilizes very strong supervision and implicitly incorporates the data from CLIP. 
Our model also achieves a competitive result compared with models pre-trained longer in epochs. 
This demonstrates that centroids as a target is a more efficient pre-training objective for classification than token-only and pixel-only.

Furthermore, we evaluate the scaling ability of our model in Table~\ref{table:vit-l} on larger ViT backbones and longer pre-training epochs. Specifically, our ViT-L model achieves 86.0\% in ImageNet-1K classification, which shows significant improvement when scaling up the ViT backbone. We outperform the MIM baseline methods BEiT by +0.8\% and MAE by +0.1\%. By pre-training longer and larger, our model achieves notable improvement, which demonstrates that our model benefits from scaling up.

\subsection{Semantic Segmentation}

Semantic segmentation aims to predict the class for each pixel, which can be considered a pixel-level classification task. 
We use ADE20k\cite{zhou2019semantic} benchmark and report the metric of mean intersection over union (mIoU) averaged over all semantic categories. About model architecture, we use ViT-Base/16 as the backbone and UPerNet\cite{xiao2018unified} as the semantic segmentation task head. For a fair comparison, we conducted fine-tuning for 160k steps and set the batch size to 16. More details of hyperparameters can be found in the Appendix.

From Table~\ref{table:results}, our method achieves 48.4 and outperforms BEiT and MAE by +3.7 and +0.3 on mIoU. We also surpass most of the methods at the same 300 epochs and fail baselines with CLIP implicitly distilled. 
The performance on ADE20K can be further improved by intermediate fine-tuning on ImageNet-1K according to BEiT. We achieve 51.6 on mIoU and gain +3.2. We exhibit the details and comparison of intermediate results in the Appendix.

\subsection{Ablation Studies}

\vspace{-20pt}

\begin{table}
    \centering
    \small
    \caption{Ablation study for different pre-training settings on ImageNet-1K classification and ADE20K segmentation. ``Replacing" means using a random replacing strategy.}
    \label{table: abla}
    \begin{tabular}{ccccc}
    \toprule
    \bf Tokens & \bf Pixel & \bf Replacing & \bf Top-1 Acc & \bf mIoU   \\
    \midrule
    \CheckmarkBold &  \XSolidBrush & \XSolidBrush & 84.12 & 47.42  \\
    \CheckmarkBold & \XSolidBrush & \CheckmarkBold &  84.21 & 47.89  \\
    \CheckmarkBold &  \CheckmarkBold & \CheckmarkBold & 84.30 & 48.35 \\
    \bottomrule
    \end{tabular}
\end{table}

We ablate the critical components, using the ViT-Base model for comparison. The evaluation was performed on image classification on ImageNet-1K and semantic segmentation on ADE20K. Pre-training was set to 300 epochs while fine-tuning consisted of 100 epochs for classification and 160k steps for segmentation.

Results are exhibited in Table~\ref{table: abla}. Regarding pre-training targets, we found that using both tokens and pixels as targets yielded the best results compared to using only tokens or pixels (84.30 vs 84.21 and 48.35 vs 47.89). Learning both token and pixel representations proved beneficial in centroid-based modeling. Additionally, we ablated different corrupting methods and observed that our model with random replacing outperformed the one with only blockwise masking in both classification (84.21 vs 84.12) and segmentation tasks (47.89 vs 47.42). This demonstrates that employing random replacement encourages the model to learn the alignment between pixels and tokens.

\subsection{Further Analysis on Centroid-based Tokenizer}

\vspace{-20pt}

\begin{table}
    \small
    \centering
    \caption{Comparison of the ratio of unchanged tokens before and after masking, and token prediction accuracy in pre-training.}\label{table: cmp_tokenizer}
    \begin{tabular}{lcccc}
    \toprule
    \multirow{2}{*}{\textbf{Method}} &   \multicolumn{3}{c}{\textbf{Mask}}  & \multirow{2}{*}{\textbf{Token Pred Acc}} \\
    \cmidrule(lr){2-4}  &  0.1  & 0.2 & 0.5  &   \\
    \midrule
    BEiT          &  34.34           & 14.17           & 1.41       & 9.49      \\
    BEiTv2            & 59.61           & 33.56           & 3.97 & 4.96  \\
    \midrule
    \textbf{Ours}     & \textbf{ 90.01} & \textbf{ 80.02} & \textbf{ 50.05}   & \textbf{40.83} \\
    \bottomrule
    \end{tabular}
\end{table}

Performances on downstream tasks are closely related to the tokenizer. Tokenizers using larger datasets (such as DALL-E in BEiT) and distilled from huge models (such as CLIP in BEiTv2) tend to achieve better results. 
However, they require a significant amount of training resources before pre-training, and take longer to infer visual tokens.

Most importantly, each visual token from these parametric tokenizers cannot accurately represent the corresponding image patch. A single token may be associated with multiple patches instead of specifically representing the patch in the same position. 
Even if a certain patch remains unchanged, the token can change if other patches are modified. This leads to misalignment between tokens and patches. 
However, the pre-training objective of previous token-based MIM methods with a mask-then-predict framework contradicts this, which aims at, predicting the masked part of visual tokens based on the masked images. Additionally, actual image patches usually exhibit local invariance, meaning that replacing one patch does not affect other patches. While existing visual tokenizers fail to maintain this local perspective, tokens from parametric tokenizers cannot independently represent single semantic information.

The experiments conducted in Table~\ref{table: cmp_tokenizer} provide evidence to support our claims. 
We start by masking out some proportion of patches and compare the output tokens with those obtained from the original, unmasked images. 
Our centroid tokenizer successfully maintains the local invariance of image patches, as the ratio of token id changes before and after the mask operation remains consistent with the ratio of the mask. 
In contrast, tokens from BEiT and BEiTv2 exhibit a higher degree of change compared to the masking ratio. Only 1.41\% and 3.97\% of tokens remain unchanged when 50\% of the tokens are masked. 
This indicates that the local correspondence between image patches and visual tokens cannot be guaranteed. 

Additionally, our model achieves the highest token prediction accuracy during pre-training, which further validates that our model is more suitable to the mask-then-predict framework and can predict visual tokens based on visual context. 
In contrast, BEiT and BEiTv2 fail to learn sufficient inter-patch relations to perform masking inference through inter-patch correlations.

\section{Conclusion and Future Work}

Existing token-based MIM with parametric tokenizer models suffers from the training cost, while pixel-based MIM needs a redundant decoder to align with the vanilla pixels. In this work, we propose a novel centroid-centered ViT pre-training framework, which utilizes centroid-based MIM to model images. 
With the k-means clustering algorithm, our approach is efficient, requiring minimal training data and time for tokenizer construction.
Our CCViT demonstrates excellent performance on downstream tasks such as classification and segmentation. 
Further analysis shows that our centroid-based tokenizer is more suitable for context-dependent architecture pre-training, due to the property of local invariant.
In the future, we plan to explore distillation approaches for improved results due to it being orthogonal to our contribution.

%
%
%
\bibliographystyle{splncs04}
\bibliography{mybibliography}

\clearpage

\appendix

\title{Appendix for CCViT}

\section{Hyperparameters for Pre-training}\label{sec:hyparams_pt}
\begin{table}[ht]
    \centering
    \small
    \caption{Hyperparameters for CCViT pre-training on ImageNet-1K.}\label{table:hype2}
    \begin{tabular}{l|c}
    \toprule
    \bf Hyperparameters         & \bf Values         \\
    \midrule
    token block layers          & 12                 \\
    pixel block layers          & 2                  \\
    layer scale                 & 0.1                \\
    patch size                  & $16 \times 16$      \\
    positional embeddings       & relative           \\
    \midrule
    pre-training epochs         & 300                \\
    batch size                  & 2048               \\
    optimizer                   & AdamW\\  
    optimizer momentum          & $\beta_1=0.9, \beta_2=0.98$ \\
    peak learning rate          & 1.5e-3             \\
    minimal learning rate       & 1e-5               \\
    learning rate schedule      & cosine delay       \\
    warmup epochs               & 10                 \\
    \midrule
    gradient clipping           & 3.0                \\
    dropout                     & \XSolidBrush       \\
    drop path                   & \XSolidBrush       \\
    stochastic path             & \XSolidBrush       \\
    weight decay                & 0.05               \\
    \midrule
    augmentation                & RandomResizedAndCrop  \\
    image resolution            & $224 \times 224$ \\
    \bottomrule
    \end{tabular}
\end{table}

\newpage

\section{Hyperparameters for Image Classification Fine-tuning}\label{sec:hyparams_ft}
\begin{table}
    \centering
    \small
    \caption{Hyperparameters for CCViT image classification fine-tuning on ImageNet-1K.}
    \begin{tabular}{l|c}
    \toprule
    \bf Hyperparameters         & \bf Values              \\
    \midrule
    positional embeddings          & relative    \\   
    \midrule
    fine-tuning epochs         & 100                \\
    batch size                  & 1024               \\
    optimizer                   & AdamW              \\  
    optimizer momentum          & $\beta_1=0.9, \beta_2=0.999$ \\
    peak learning rate          & 2e-3             \\
    minimal learning rate          & 1e-6             \\
    learning rate schedule      & cosine delay       \\
    layer-wise learning rate decay & 0.65             \\
    warmup epochs               & 20                 \\
    \midrule
    gradient clipping              & \XSolidBrush     \\
    label smoothing                & 0.1              \\
    dropout                        & \XSolidBrush     \\
    stochastic depth               & 0.1              \\
    weight decay                   & 0.05             \\
    \midrule 
    augmentation                   & RandAug (9, 0.5)  \\
    image resolution               & $224 \times 224$ \\
    random erasing                 & 0.25             \\
    mixup                          & 0.8              \\
    cutmix                         & 1.0              \\
    \bottomrule
    \end{tabular}
\end{table}

\newpage

\section{Hyperparameters for Semantic Segmentation Fine-tuning}\label{sec:hyparams_seg}

\vspace{-30pt}
\begin{table}[ht]
    \centering
    \small
    \caption{Hyperparameters for CCViT pre-training on ImageNet-1K.}\label{table:hype1}
    \begin{tabular}{l|c}
    \toprule
    \bf Hyperparameters         & \bf Values         \\
    \midrule
    positional embeddings          & relative    \\   
    \midrule
    fine-tuning steps & 160K \\
    batch size & 16 \\
    optimizer                   & AdamW              \\  
    optimizer momentum             & $\beta_1=0.9, \beta_2=0.999$ \\
    peak learning rate & 8e-5 \\
    minimal learning rate & 0 \\
    learning rate schedule         & linear                       \\
    layer-wise learning rate decay & 0.9                          \\
    warmup steps & 1500 \\
    \midrule
    dropout & \XSolidBrush \\
    stochastic depth & 0.1 \\
    weight decay & 0.05 \\
        \midrule
    input resolution & $512 \times 512$ \\
    \bottomrule
    \end{tabular}
\end{table}
\vspace{-30pt}

\section{Ablation Study of Centroid Tokenizer}
\vspace{-20pt}

\begin{figure*}
    \begin{center}
       \includegraphics[width=0.9\linewidth]{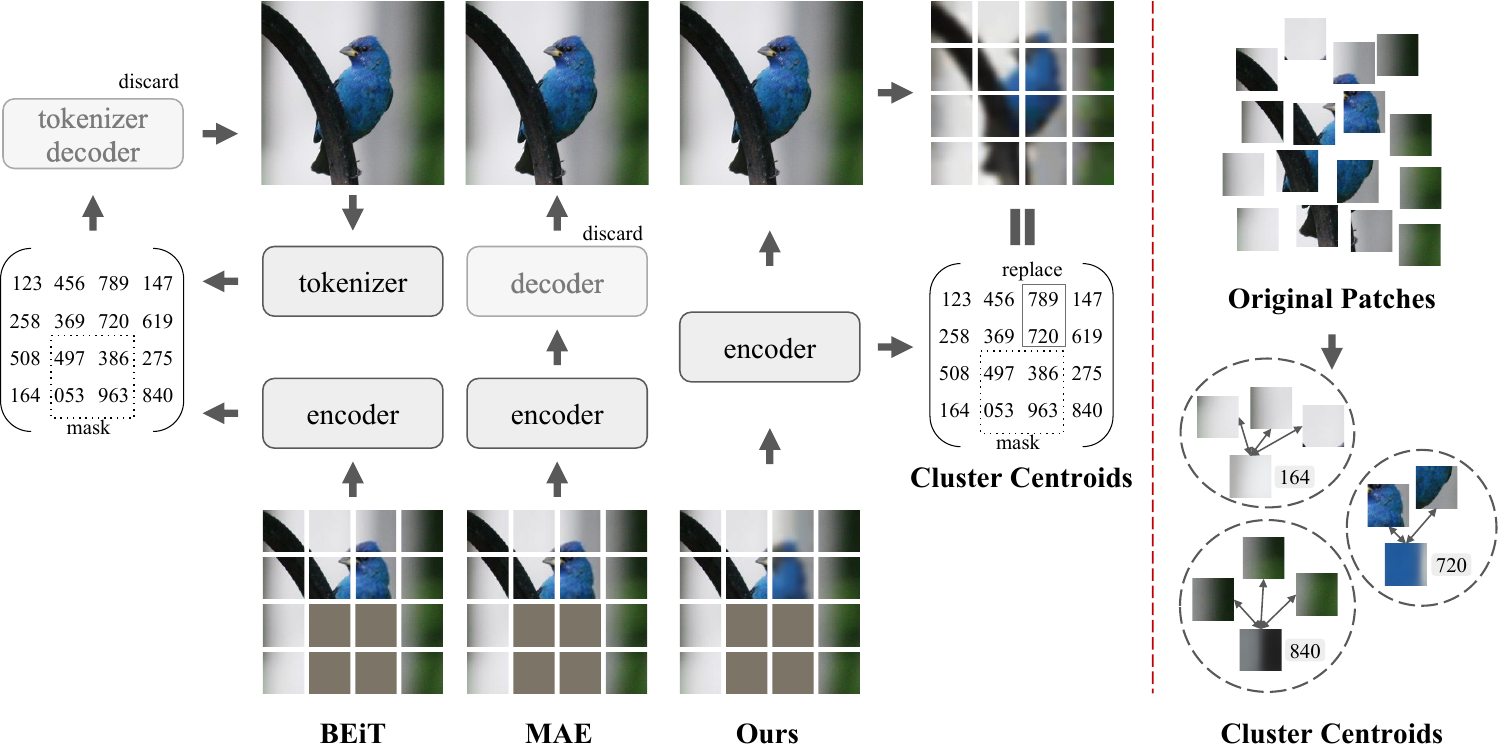}
    \end{center}
\vspace{-15pt}
   \caption{Comparison of pre-training architectures between BEiT, MAE, and ours.}
\label{fig:cmp}
\vspace{-15pt}
\end{figure*}

Figure~\ref{fig:cmp} shows the main comparison of pre-training architectures between our centroid-based CCViT, token-based BEiT, and pixel-based MAE. Our input consists of both masked patches, replaced patches (centroids), and original patches. Our model architecture does not feature a redundant decoder, further streamlining the process. We use both tokens and pixels as our pre-training objectives.

\clearpage

\section{Ablation Study of Centroid Tokenizer}

\vspace{-20pt}

\begin{table}
    \centering
    \small
    \caption{Fine-tuning results of ImageNet-1K Image Classification under BEiT ``framework". $^\dag$: It is unfair to compare baselines with BEiTv2, as it employs an additional ViT branch of patch aggregation. Also, its tokenizer is obtained by distilling from strong CLIP supervisions.}\label{table:beit}
    \label{table-ef}
    \begin{tabular}{lccc}
    \toprule
    \bf Method     & \bf Epoch & \bf ViT-B & \bf ViT-L        \\
    \midrule
    BEiT         & 800       & 83.2  &  85.2 \\
    BEiTv2$^\dag$       & 1600       & 85.5  &  87.3 \\
    mc-BEiT        & 800       & 84.1  &  85.6 \\
    MAE        & 400     & 83.1  & 84.5  \\
    \midrule
    \bf Ours     & \bf 300  &  \bf 84.2 &  \bf 86.0    \\
    \bottomrule
    \end{tabular}
\end{table}

In this setting, we further ablate the effectiveness of the visual tokenizer. We report different baselines on an overall BEiT-style framework. For our CCViT, we remove the Pixel ViT block and only use tokens as pre-training objectives to match the BEiT framework. For MAE, as its pre-training objectives are based on pixels, we report the result from an implementation based on BEiT framework~\cite{github}. \footnote{GitHub Implementation of MAE based on BEiT framework: \href{https://github.com/pengzhiliang/MAE-pytorch}{https://github.com/pengzhiliang/MAE-pytorch}.} Results are reported in Table~\ref{table-ef}. Our tokenizer is the most effective method compared with other visual tokenizers. Note that BEiTv2 is supervised by strong CLIP models and employs different architecture, thus is unfair to compare directly.

\section{Intermediate Fine-tuning on ADE20K}

\vspace{-20pt}

\begin{table}
    \centering
    \small
    \caption{Intermediate Fine-tuning on ADE20k, which is pre-trained and fine-tuned on ImageNet-1K classification.}\label{table:inter}
    \begin{tabular}{lccc}
    \toprule
    \bf Method    & \bf Supervision & \bf Epoch & \bf mIoU   \\
    \midrule
    BEiT      & DALL-E   & 800    & 45.6   \\
    mc-BEiT   & VQGAN       & 800  & 47.0   \\
    Ours      &  Centroids & 300  &  48.4    \\
    \multicolumn{4}{l}{\textit{+ Intermediate fine-tuning}} \\
    BEiT     & DALL-E   & 800     & 47.7 \\
    mc-BEiT  & VQGAN       & 800    & 50.8 \\
    \bf Ours      &  Centroids & \bf 300  &  \bf 51.6    \\
    \bottomrule
    \end{tabular}
\end{table}

According to BEiT, the performance on ADE20K can be further improved by intermediate fine-tuning on ImageNet-1K. So we conducted this experiment and compared it with BEiT and mc-BEiT. We first immediately fine-tune on ImageNet-1K for 100 epochs and then fine-tune on ADE20K semantic segmentation normally. In Table~\ref{table:inter} we report the performance results. Our CCViT method achieves 51.6\% on mIoU, and gains +3.2\% to our pre-training-only model, 
and significantly outperforms prior arts BEiT and mc-BEiT, with the +3.9\% and +0.8\% on mIoU respectively.

\section{Performance of Different Tokenizers}\label{sec:per_tokenizers}

\vspace{-20pt}

\begin{table}[ht]
    \centering
    \small
    \caption{Performance of different tokenizers.}\label{table: performance}
    \begin{tabular}{lccccc}
    \toprule
    \bf Method & \textbf{Train} & \textbf{Inference} & \textbf{Mem.}  & \textbf{Params} & \textbf{FLOPs} \\
    \midrule
    BEiT &  Days    &   15.7$\pm$2.5ms   & 21.5G  &   98M &   3.3T   \\
    BEiTv2  & Days &  74.8$\pm$0.6ms   &  13.2G  &  109M  &   2.3T  \\
    \midrule
    \bf Ours   &   \bf 158s   & \bf  9.3$\bm{\pm}$0.5ms &  \bf 3.4G & \bf 0 & \bf 805M  \\
    \bottomrule
    \end{tabular}
\end{table}

\vspace{-20pt}

\begin{table}[ht]
    \small
    \centering
    \caption{Comparison of the ratio of unchanged tokens under different image noises.}\label{table: cmp_2}
    \begin{tabular}{lcccccc}
    \toprule
    \multirow{2}{*}{\textbf{Method}} & \multicolumn{3}{c}{\textbf{Gaussian Noise}}  &  \multicolumn{3}{c}{\textbf{Gaussian Blur}}  \\
    \cmidrule(lr){2-4} \cmidrule(lr){5-7}  & 1 & 10 & 25  & 0.5   & 1    & 2   \\
    \midrule
    BEiT                 & 88.02           & 32.54           & 9.31            &  61.18           & 25.32           & 6.93        \\
    BEiTv2         & 95.03           & 57.43           & 24.02           & 83.52           & 61.29           & 0.08          \\
    \midrule
    \textbf{Ours}    & \textbf{ 98.94} & \textbf{ 88.61} & \textbf{ 72.28}     & \textbf{ 96.38} & \textbf{ 86.39} & \textbf{ 66.72} \\
    \bottomrule
    \end{tabular}
\end{table}

In this section, we show the performance of different tokenizers in Table~\ref{table: performance} and Table~\ref{table: cmp_2}. 

For the tokenizer training speed, the centroid-based tokenizer only needs about 158s on a single RTX 3090 NVIDIA GPU to construct the tokenizer, while tokenizers in BEiT and BEiTv2 need several days\footnote{It depends on the hardware environments. BEiTv2 tokenizer costs 7 days of training on a single RTX 3090 NVIDIA GPU.}. 
For the tokenizer inference speed, under the same batch size consisting of 64 images, and the same environment (a single RTX 3090), our model only uses 9.3ms, while BEiT and BEiTv2 tokenizers use 15.7 and 74.8ms respectively. It is worth noting that our tokenizer occupies very little GPU memory, and our speed can be scaled up to a larger batch size to fully utilize the parallel capability of the GPU, which will further amplify our speed advantage. 
We further calculate the FLOPs of different tokenizers, and our centroid tokenizer outperforms others in large margins, attributed to the non-parametric design.

\begin{figure}[ht]
    \begin{center}
       \includegraphics[width=0.85\linewidth]{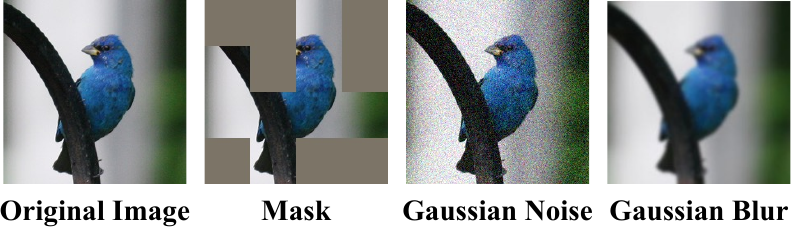}
    \end{center}
    \caption{Visualization of different image noises in Table~\ref{table: cmp_tokenizer} and Table~\ref{table: cmp_2}.}
    \label{fig:stable}
\end{figure}

We evaluate the noise resistance ability of different tokenizers under different image noises. We exhibit the results by measuring the ratio of unchanged tokens in Table~\ref{table: cmp_2}. Our centroid tokenizer consistently outperforms other parametric tokenizers by a significant margin, especially when higher degrees of noise are applied. Additionally, Figure~\ref{fig:stable} visualizes the different noise types explored.

\section{Pixel Reconstruction}\label{sec:pixel_reconstruct}

In this section, we demonstrate that our CCViT has the ability of semantic predictions via visual context. We explore the plausibility of utilizing solely a pre-trained encoder for image reconstruction without any additional models. Previous pixel-based MIM uses a decoder and raw pixels as pre-training targets to predict masked patches, while token-based MIM needs additional tokenizer models. Only utilizing the centroids and token probability vector output from the encoder, our CCViT can easily reconstruct the masked part. Note that MIM methods are not designed to achieve satisfactory results in image inpainting tasks.

\begin{figure}[ht]
    \begin{center}
       \includegraphics[width=0.7\linewidth]{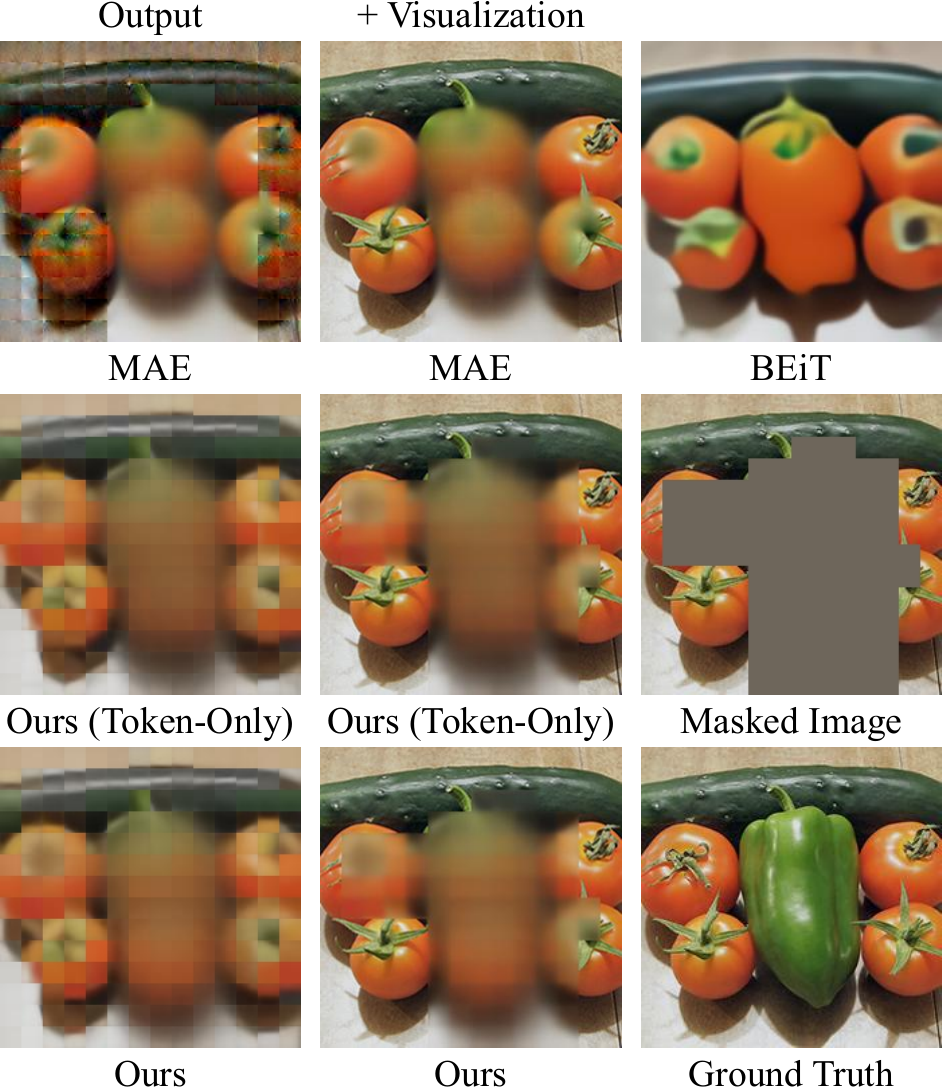}
    \end{center}
    \caption{Reconstruction examples from ImageNet-1k \textit{validation} dataset via BEiT, MAE and our CCViT.}
    \label{fig:inpainting}
\end{figure}

We compare the reconstruction examples of BEiT~\cite{bao2022beit}, MAE~\cite{he2022masked}, and our CCViT, which stands for token-based, pixel-based, and centroid-based MIM respectively. Pixel-based MAE utilizes the direct output from pre-trained models as its pre-training targets are raw pixels. Token-based BEiT utilizes an additional tokenizer decoder with the predicted masked tokens as input. Our centroid-based CCViT directly utilizes the corresponding centroids of the predicted masked tokens, thereby avoiding the need for additional parametric models. We show results from both the token-only version and full version CCViT, \ie\ only index tokens as pre-training targets and both tokens and pixels as targets respectively. Reconstruction examples are shown in Figure~\ref{fig:inpainting}. As MAE tends to destroy the unmasked patches and to alleviate the problem of obvious borderlines between patches, we replace the unmasked patches with the ground truth and add Gaussian blur operation on our CCViT output for visualization. 

\begin{figure*}
    \begin{center}
       \includegraphics[width=0.95\linewidth]{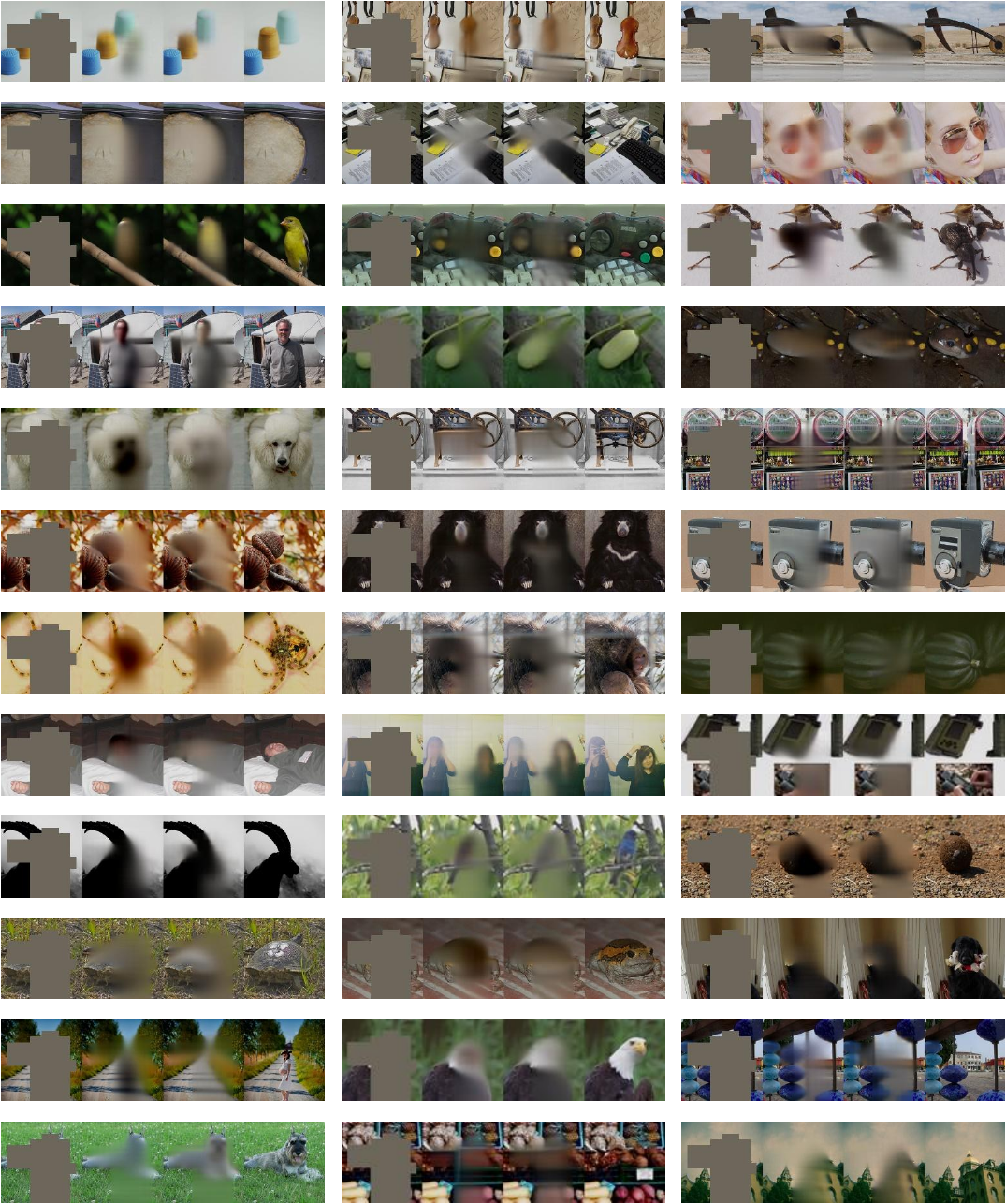}
    \end{center}
    \caption{Example samples of image reconstruction on ImageNet-1K \textit{test} images. For each quadruplets, we show the masked image (1st from left), MAE reconstruction (2nd from left), our CCViT reconstruction (2nd from right), and the ground truth (1st from right). The masking ratio is 50\%.}
    \label{fig:inpainting_large}
\end{figure*}

In this case, most of the ``green pepper" in the image is masked and only the stem remains visible. Reconstruction from BEiT is far from the ground truth. MAE simply pastes the surrounding images and fills in two oranges under the stem. Our CCViT performs better in shape prediction. Token-only CCViT even predicts better. This suggests that despite having strictly pre-trained solely on centroid token targets, and neglecting any explicit introduction of pixel targets, CCViT can still implicitly predict the pixels via its centroid tokens, and learn the correlation between centroid pixels and indices. More reconstruction examples are shown in Figure~\ref{fig:inpainting_large}.

\section{Visualization of centroids}
\begin{figure*}
    \begin{center}
       \includegraphics[width=0.95\linewidth]{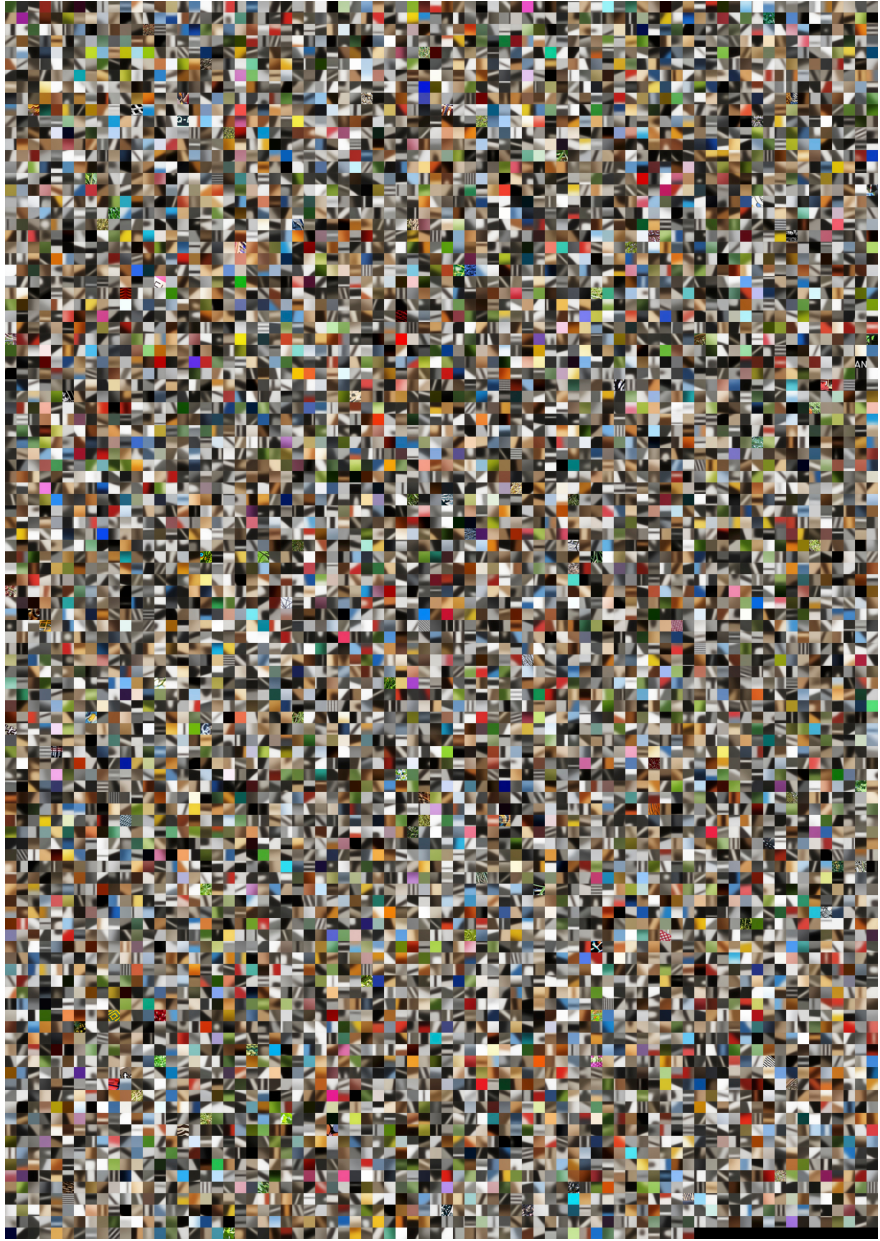}
    \end{center}
    \caption{Visualization of 8192 centroids.}
    \label{fig:centroids}
\end{figure*}

Different from token-based MIM which makes it difficult to explicitly visualize the single semantics of each token, we show all of our 8192 centroids in Figure~\ref{fig:centroids}. Each centroid is a $16\times16$ image patch, which also stands for an index token.

\end{document}